%% file: main.tex
\documentclass[letterpaper]{article} 
\usepackage[draft]{aaai25}  
\usepackage{times}  
\usepackage{helvet}  
\usepackage{courier}  
\usepackage[hyphens]{url}  
\usepackage{graphicx} 
\usepackage{natbib}  
\usepackage{caption} 
\usepackage{algorithm}
\usepackage{algorithmic}
\usepackage{float}

\usepackage{newfloat}
\usepackage{listings}
\DeclareCaptionStyle{ruled}{labelfont=normalfont,labelsep=colon,strut=off} 
\floatstyle{ruled}
\newfloat{listing}{tb}{lst}{}
\floatname{listing}{Listing}
\usepackage{multirow}
\usepackage{xcolor}
\usepackage{hhline}
\usepackage{colortbl}
\usepackage{booktabs}
\usepackage{tabularx}
\usepackage{makecell}
\usepackage{subcaption}
\usepackage{lipsum}
\usepackage{caption}
\usepackage{amsmath}
\usepackage{amsfonts}
\usepackage{graphicx}
\usepackage{float}
\usepackage{array}

\usepackage{color}
\usepackage{hyperref}
\hypersetup{
}

\usepackage{xspace}

\def\tech{\textit{RILQ}\xspace}

\title{RILQ: \underline{R}ank-\underline{I}nsensitive \underline{L}oRA-based \underline{Q}uantization Error Compensation \\ for Boosting 2-bit Large Language Model Accuracy}

\author{Geonho Lee${}^{1}$\thanks{\, Equal contribution\quad${}^\dagger$Corresponding author},
        Janghwan Lee${}^{1}$\footnotemark[1],
        Sukjin Hong${}^{2}$\footnotemark[1],
        Minsoo Kim${}^{1}$, \\
        Euijai Ahn${}^{2}$,
        Du-Seong Chang${}^{3}$ and 
        Jungwook Choi${}^{1\dagger}$}
\affiliations{
        \normalsize{\textsuperscript{1}Hanyang University},
        \normalsize{\textsuperscript{2}KT}, 
        \normalsize{\textsuperscript{3}Sogang University} \\
        \normalsize{Seoul, Republic of Korea}\\
        \small{\textsuperscript{1}\texttt{\{thisisho, hwanii0288, minsoo2333, choij\}@hanyang.ac.kr}} \\
        \small{\textsuperscript{2}\texttt{\{sukjin.hong, euijai.ahn\}@kt.com}},  \small{\textsuperscript{3}\texttt
        {dschang@sogang.ac.kr}} \\   
}

\usepackage{bibentry}

\begin{document}

\maketitle

\begin{abstract}
Low-rank adaptation (LoRA) has become the dominant method for parameter-efficient LLM fine-tuning, with LoRA-based quantization error compensation (LQEC) emerging as a powerful tool for recovering accuracy in compressed LLMs. However, LQEC has underperformed in sub-4-bit scenarios, with no prior investigation into understanding this limitation. We propose \tech (\underline{R}ank-\underline{I}nsensitive \underline{L}oRA-based \underline{Q}uantization Error Compensation) to understand fundamental limitation and boost 2-bit LLM accuracy. Based on rank analysis revealing model-wise activation discrepancy loss's rank-insensitive nature, \tech employs this loss to adjust adapters cooperatively across layers, enabling robust error compensation with low-rank adapters. Evaluations on LLaMA-2 and LLaMA-3 demonstrate \tech's consistent improvements in 2-bit quantized inference across various state-of-the-art quantizers and enhanced accuracy in task-specific fine-tuning. \tech maintains computational efficiency comparable to existing LoRA methods, enabling adapter-merged weight-quantized LLM inference with significantly enhanced accuracy, making it a promising approach for boosting 2-bit LLM performance.\footnote{Our code is available at \href{https://github.com/aiha-lab/RILQ}{\textcolor{blue}{https://github.com/aiha-lab/RILQ}}}
\end{abstract}
\vspace{-2.5mm}

\input{Sections/introduction}
\input{Sections/relatedwork}
\input{Sections/background}
\input{Sections/methodology}
\input{Sections/experiments}
\input{Sections/conclusion}

\section{Acknowledgments}
This work was partly supported by Institute of Information \& communications Technology Planning \& Evaluation (IITP) grant funded by the Korea government (MSIT) (No. RS-2020-II201373, Artificial Intelligence Graduate School Program (Hanyang University), and 2022-0-00957, Distributed on-chip memory-processor model PIM (Processor in Memory) semiconductor technology development for edge applications) and National Research Foundation of Korea (NRF) (No. RS-2023-00260527). 

\bibliography{aaai25}

\clearpage
\input{Sections/appendix}

\end{document}

%% file: Sections/introduction.tex
\section{Introduction}
\label{sec:introduction}

\begin{figure}[t]
\begin{center}
\centerline{\includegraphics[width=\columnwidth]{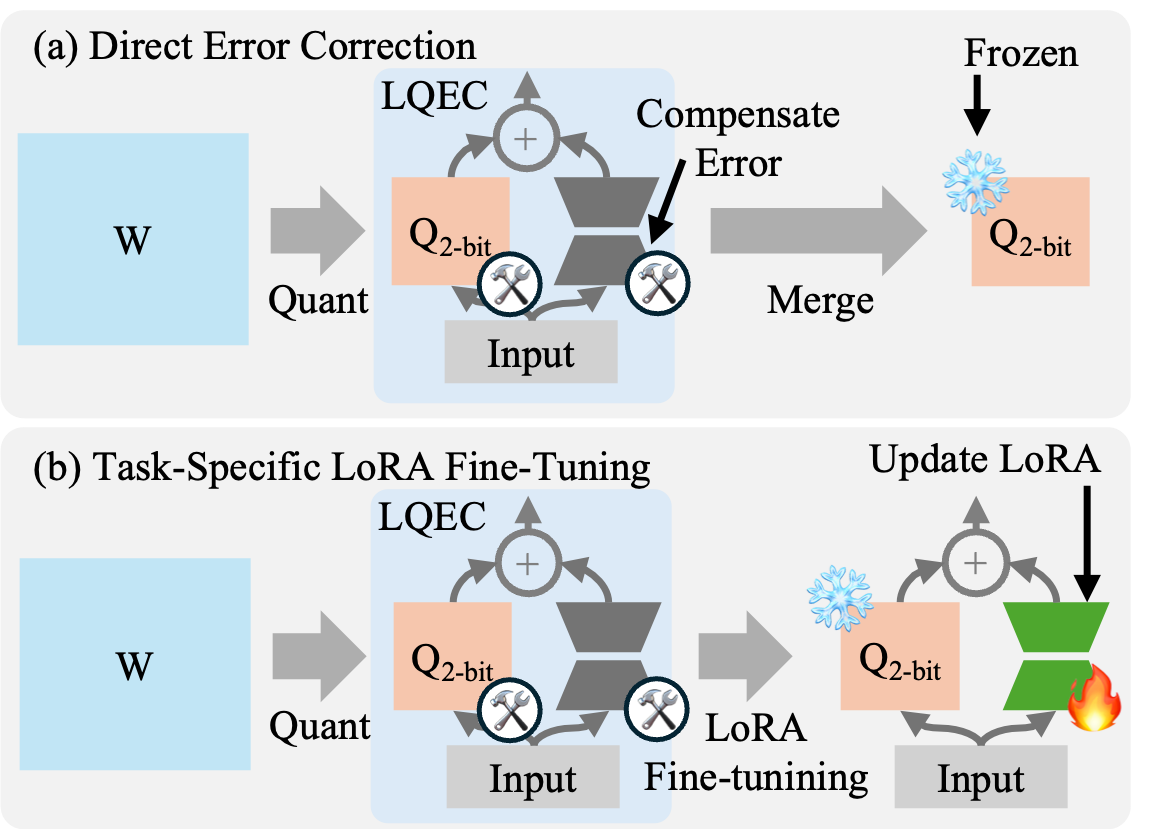}}
\caption{LoRA-based quantization error compensation (LQEC): (a) direct error correction, (b) initialization for task-specific fine-tuning.}
\vspace{-5mm}
\label{fig:overview}
\end{center}
\end{figure}

Large language models (LLMs) like GPT-4~\cite{gpt4} and LLaMA-3~\cite{llama3modelcard} have revolutionized various domains, demonstrating human-level performance in complex tasks such as question answering~\cite{kamalloo-etal-2023-evaluating}, code auto-completion~\cite{rozière2024code}, and summarization~\cite{10.1162/tacl_a_00632_news}. However, adapting these models to specialized domains requires efficient fine-tuning techniques~\cite{wei2022finetuned,wang-etal-2023-self-instruct}. \textit{Low-rank adaptation} (LoRA)~\cite{2021.Hu.Chen} has emerged as a leading solution, efficiently reparameterizing weight matrices with low-rank adapters to incorporate task-specific information. By fine-tuning only a small, adaptable extension to the frozen base model, LoRA significantly reduces the memory footprint while effectively specializing LLMs. This cost-effective fine-tuning approach has expanded to enable autonomous module composition~\cite{2023.Huang.Lin}, multiple-task adaptation~\cite{2023.Sheng.Stoica}, and long-context inference~\cite{2023.Chen.Jia}. LoRA has emerged as a powerful tool for quantization error compensation in large language models (LLMs), addressing the challenges of reduced inference and fine-tuning costs~\cite{2023.Dettmers.Zettlemoyer}. While weight quantization techniques~\cite{2024.Ashkboos.Hensman,2023.Frantar.Alistarh,2023.Lee.Lee,2022.Yao.He,2023.Lin.Han} mitigate LLMs' memory footprint, they introduce errors due to reduced-precision representation. Therefore, LoRA has been adopted to help compensate quantization error. Fig.~\ref{fig:overview} illustrates two LoRA-based quantization error compensation (LQEC) approaches: (1) Direct error correction, where adapters offset quantization errors in weights. For instance, ZeroQuant-v2~\cite{2023.Yao.Heouq} employs a low-rank adapter per linear module, which can be merged into quantized weights~\cite{xu2024qalora,liu2024qllmaccurateefficientlowbitwidth} for efficient inference (Fig.~\ref{fig:overview}(a)). (2) Task-specific fine-tuning, exemplified by LoftQ~\cite{2023.Guo.Kim}, which combines LoRA with the base model quantized and frozen for memory-efficient fine-tuning, using LQEC-tuned adapters as initialization (Fig.~\ref{fig:overview}(b)). Notably, LQEC integrates into existing structures without requiring additional adapters, offering a promising method to address LLM's memory bottleneck while preserving accuracy.

Despite advances in LQEC, achieving high compression rates, such as 2-bit quantization, without compromising model accuracy remains challenging. Aggressive quantization, particularly at sub-4-bit precision, often leads to significant accuracy degradation. Various initiatives have attempted to bridge this accuracy gap by minimizing quantization-induced discrepancies. Weight-level approaches like LQ-LoRA~\cite{2023.Guo.Kim}, LoftQ~\cite{2023.Li.Zhao}, LQER~\cite{zhang2024lqer}, RA-LoRA~\cite{ralora}, and RoLoRA~\cite{huang2024rolorafinetuningrotatedoutlierfree} address discrepancies at each weight quantization by factorizing errors into low-rank adapters via singular value decomposition (SVD). ApiQ~\cite{2024.Liao.Monz} compensates for quantization error via gradient updates based on losses at each linear module's output, while QLLM~\cite{liu2024qllmaccurateefficientlowbitwidth} and ReALLM~\cite{leconte2024reallmgeneralframeworkllm} use losses at the output of Transformer decoder layers. While these approaches have partially mitigated quantization errors through various ad-hoc techniques, they still suffer significant accuracy losses at 2-bit levels. Moreover, there remains a lack of fundamental understanding as to why LQEC struggles to compensate for aggressive bit-precision quantization, highlighting the need for in-depth analysis.

In this work, we explore the challenges of LQEC under 2-bit weight quantization, highlighting that \textit{lower bit-precision necessitates higher ranks for effective error compensation}, which contradicts LoRA's low-rank premise. We introduce rank sensitivity analysis to assess the rank requirements for error compensation and find that \textit{rank sensitivity decreases as discrepancy scope increases}. Leveraging this insight, we propose \tech (\underline{R}ank-\underline{I}nsensitive \underline{L}oRA-based \underline{Q}uantization Error Compensation), which employs a model-wise discrepancy loss at the output of the last Transformer layer. This approach enables cooperative adjustment of both rank-redundant and rank-critical linear modules during LoRA tuning, facilitating flexible signal propagation and quantization error compensation at the model output. Our method delivers robust LQEC even with small ranks (e.g., 16-rank), recovering accuracy while preserving the efficiency of existing LQEC techniques. Evaluations on LLaMA-2 and LLaMA-3 demonstrate consistent improvements in 2-bit quantized inference across state-of-the-art quantizers (OmniQuant~\cite{2023.Shao.Luo}, QuIP~\cite{2023.Chee.Sa,2024.Tseng.Sa}, QuaRot~\cite{ashkboos2024quarot}) and enhanced accuracy in task-specific fine-tuning without additional inference cost, enabling efficient adapter-merged weight-quantized LLM inference of  QA-LoRA~\cite{xu2024qalora} with significant accuracy boosts. These results suggest that our method effectively repositions LQEC as a promising accuracy enhancer for 2-bit LLM inference.

%% file: Sections/relatedwork.tex
\begin{figure*}[t]
\begin{center}
\centerline{\includegraphics[width=0.9\textwidth]{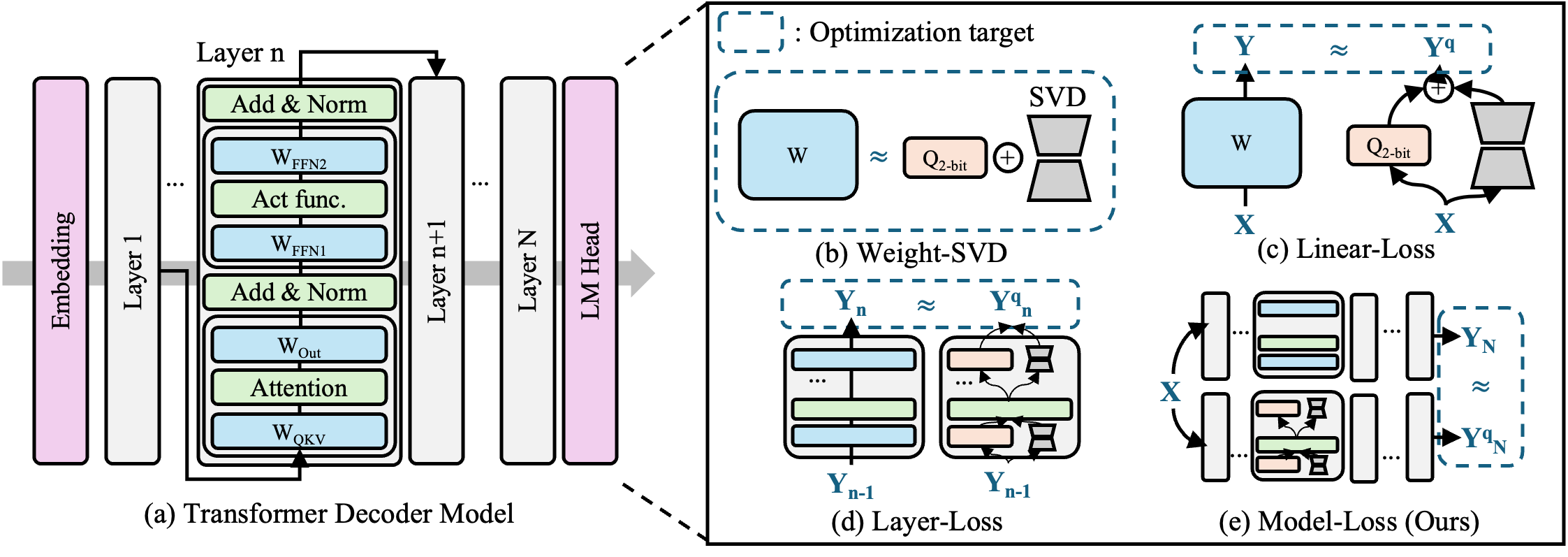}}
\caption{(a) Structure of the Transformer decoder model. (b-e) Four optimization approaches for fine-tuning LoRA for quantization error compensation.}
\label{fig:optimization}
\vspace{-8mm}
\end{center}
\end{figure*}

\section{Related Work}
\label{sec:relatedwork}

\textbf{LLM Weight Quantization. }
Weight quantization is a promising technique to reduce the memory footprint of LLMs by lowering the bit-precision of weight values~\cite{2023.Frantar.Alistarh,2023.Lee.Lee,2022.Yao.He,2023.Lin.Han,2022.Wei.Liu, 2023.Wei.Liu}. Activation-aware weight quantization methods~\cite{2023.Lin.Han, 2024.Lee.Park, 2023.Kim.Keutzer, 2023.Guo.Zhu, 2023.Heo.Lee} have successfully reduced weight precision to 4-bit or lower, but these approaches incur mixed-precision overhead and struggle with poor accuracy at 2-bit precision. Parallel efforts have focused on developing advanced quantizers for 2-bit LLMs~\cite{2022.Yao.He, 2023.Shao.Luo, 2023.Chee.Sa, 2024.Tseng.Sa, 2024.Guan.Yu, 2024.Egiazarian.Alistarh}. For instance, OmniQuant~\cite{2023.Shao.Luo} introduced learnable quantization parameters to adjust weight ranges and transformations, QuIP~\cite{2023.Chee.Sa, 2024.Tseng.Sa} employed grouped vector quantization with a shared codebook for non-uniform weight group representation, and QuaRot~\cite{ashkboos2024quarot} rotated the weight matrix to redistribute outliers. While these non-uniform quantization schemes achieve robust 3$\sim$4-bit LLM inference accuracy, they still face challenges in closing the accuracy gap for 2-bit LLM inference.

\textbf{LoRA for Fine-tuning and Compensation. }
Low-rank adaptation (LoRA)\cite{2021.Hu.Chen} has emerged as the leading parameter-efficient fine-tuning technique to enhance the capabilities of foundational large language models (LLMs)~\cite{2021.Hu.Chen,2023.Huang.Lin,2024.Xia.Hazan, 2023.Hu.Poria,2022.Ding.Sun,2024.Han.Zhang, 2023.Chen.Yang,2023.Sheng.Stoica}. LoraHub~\cite{2023.Huang.Lin} aggregates LoRA modules trained on different tasks to autonomously compose compatible modules, while SLoRA~\cite{2023.Sheng.Stoica} facilitates multiple-LoRA blocks for various tasks, and LongLoRA~\cite{2023.Chen.Jia} enables context extension for long-context LLM inference. LoRA has also been adapted for \textit{error compensation}, with ZeroQuant-v2~\cite{2023.Yao.Heouq} employing a low-rank adapter to compensate for weight quantization errors and similar approaches used for pruning error compensation~\cite{2024.Li.Zhang,2023.Zhang.Zhuang,2023.Chen.Liang}. Despite these advances, LoRA-based error compensation methods still face challenges in achieving high compression rates, such as 2-bit quantization, without compromising model accuracy.

\textbf{Quantization Error Compensation.}
Quantization error compensation (QEC) has been extensively explored in the context of quantization-aware training (QAT), which supports aggressive sub-4bit quantization while preserving task-specific fine-tuned accuracy~\cite{2023.Kim.Choi, 2023.Liu.Chandra}. For example, TSLD~\cite{2023.Kim.Choi} uses full-parameter tuning combined with token-wise scaled knowledge distillation to enable robust 2-bit LLM inference, though it requires significant memory for full-parameter adjustments, similar to pre-training. To reduce the memory demands of QAT, LoRA-based parameter-efficient QEC have been developed~\cite{2023.Dettmers.Zettlemoyer,xu2024qalora,ralora,2023.Guo.Kim, 2023.Li.Zhao, 2024.Liao.Monz,2023.Chai.Wei}. Notably, QLoRA~\cite{2023.Dettmers.Zettlemoyer} applies LoRA on top of frozen quantized weights to minimize memory usage, while QA-LoRA~\cite{xu2024qalora} integrates adapters into quantized weights to enhance LLM inference efficiency. Despite these advancements, significant accuracy losses still occur with extremely low-bit quantization. Recent quantization-aware LoRA methods, such as LoftQ~\cite{2023.Li.Zhao} as well as \cite{2023.Guo.Kim,zhang2024lqer,ralora,huang2024rolorafinetuningrotatedoutlierfree} tackle these challenges by using singular value decomposition (SVD) to factorize quantization errors into low-rank adapters. \cite{2024.Liao.Monz,liu2024qllmaccurateefficientlowbitwidth,leconte2024reallmgeneralframeworkllm} address quantization errors through gradient updates based on losses at each linear module or Transformer decoder layer's output. Despite these efforts, maintaining accuracy at 2-bit precision remains a significant challenge with no understanding about why.

%% file: Sections/background.tex
\begin{figure}[t]
\begin{center}
\centerline{\includegraphics[width=\columnwidth]{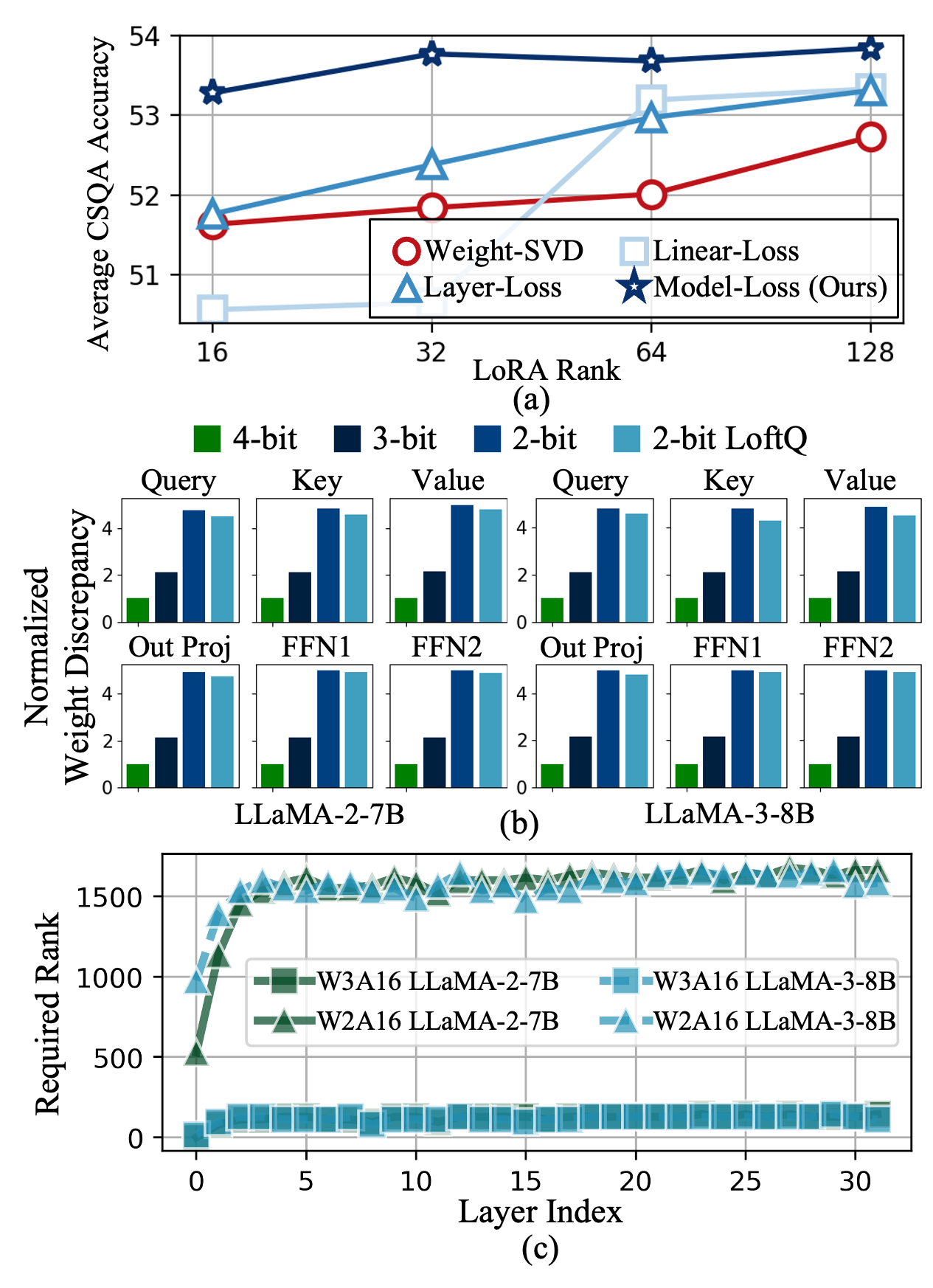}}
\vspace{-3mm}
\caption{(a) Average CSQA accuracy across optimization granularity and the rank of LoRA (LLaMA-2-7B). (b) Normalized weight discrepancy $(\|W-Q\|_{F})$ across models (LLaMA-2-7B and LLaMA-3-8B) and every linear module, normalized to 1 for 4-bit quantization discrepancy. (c) Minimum rank required for each quantization bit-precision to closely achieve the weight discrepancy of 4-bit quantization.} 
\vspace{-5mm}
\label{fig:w_distance}
\end{center}
\end{figure}

\section{Background and Challenges}
\label{sec:background}

\subsection{LoRA-based Quantization Error Compensation}
\label{subsec:quantize_and_lora}
LLMs typically comprise multiple Transformer decoder layers, an Embedding layer for token-to-activation conversion, and an LM-Head for converting Transformer outputs into vocabulary logits (Fig.\ref{fig:optimization}(a)). Each Transformer layer contains several linear modules with weight parameters ($W_{QKV},W_{Out},W_{FFN1},W_{FFN2}$) for matrix multiplication with input ($X$) to compute output activation ($Y$). Since the number of parameters exceeds billions, weight quantization, a prominent model compression technique, represents a pre-trained weight matrix $W \in \mathbb{R}^{d_1\times d_2}$ using a limited bit width $b$. The quantized weight $Q_b$ is formulated by Eq.~\ref{eq:quantizer}:

\begin{equation}
\begin{aligned}
Q_b &= s \cdot \text{clamp}(\left\lfloor \frac{W}{s}\right\rceil-z, 0, 2^N - 1) + z \\
s &= \frac{\gamma\text{max}(W) - \beta\text{min}(W)}{2^b - 1} \text{,  } z = \left\lfloor \frac{\beta\text{min}(W)}{s} \right\rceil,
\end{aligned}
\label{eq:quantizer}
\end{equation}
where $ \left\lfloor \cdot \right\rceil $ indicates the rounding function. The choice of $\gamma$ and $\beta$ varies depending on the quantizers; they can be a constant ($\gamma=1$, $\beta=1$) for the round-to-nearest quantization (RTN) or represent the learnable clipping strengths for the upper and lower bounds of quantized weights as in OmniQuant~\cite{2023.Shao.Luo}.

LoRA introduces learnable low-rank adapter modules, $L_1 \in \mathbb{R}^{d_1\times r}$ and $L_2 \in \mathbb{R}^{d_2\times r}$ ($r \ll \max(d_1, d_2)$), with frozen pre-trained $W$, formulating the forward operation for a linear module in the Transformer layer as $ Y = X(W + L_{1} {L_{2}}^T)$. For LoRA-based QEC, rank-$r$ adapters are updated to compensate for the impact of weight quantization. For example, LoftQ~\cite{2023.Li.Zhao} updates the adapters to minimize the weight discrepancy (Weight-SVD, Fig.~\ref{fig:optimization}(b)): 
\begin{equation}
\underset{L_1, L_2}{\arg\min} \ \|W - (Q_b + L_1L_2^{\top})\|_{F},
\label{eq:loftq_obj}
\end{equation}
where $Q_b =$ Quant$(W-L_1L_2^{\top})$, and $L_1L_2^{\top}$ is iteratively updated via SVD. However, weight discrepancy optimization does not consider the combined impact of $W$ and $X$ to the matrix multiplication output $Y$. Thus, \cite{2024.Liao.Monz} proposed the discrepancy optimization on the output of linear modules (Linear-Loss, Fig.~\ref{fig:optimization}(c)):
\begin{equation}
\underset{L_1, L_2}{\arg\min} \ \|Y-Y^q\|_{F},
\label{eq:apiq_obj_eq}
\end{equation}
where $Y=WX$ and $Y^q=(Q_b + L_1L_2^{\top})X$. QLLM~\cite{liu2024qllmaccurateefficientlowbitwidth} further extended the optimization scope to a Transformer layer by updating $L_1$ and $L_2$ parameters within it via gradient from the loss between $Y_n$ and $Y^q_n$ defined as :  
\begin{equation}
\begin{aligned}
Y_n &= G(Y_{n-1}, \{W_{n,l}\}_{l=1}^L) \\
~~Y^q_n &= G(Y^q_{n-1}, \{Q_{b,n,l}, L_{1,n,l},L_{2,n,l}\}_{l=1}^L), 
\label{eq:block}
\end{aligned}
\end{equation}
where $n$ is a layer index and $G$ represents a group of $L$ linear modules within a Transformer layer which are sequentially processed according to the Transformer structure. As described in Fig.~\ref{fig:optimization}(d), this layer-wise discrepancy loss (Layer-Loss) asserts the alignment of quantized output activation to the FP16 output at each Transformer layer.

\subsection{LQEC's Challenge for 2-bit LLM}
Despite efforts to compensate for quantization errors at various scopes, LQEC has shown limited success with aggressive sub-4-bit quantization. Fig.~\ref{fig:w_distance}(a) demonstrates that existing discrepancy minimization approaches (Weight-SVD, Linear-Loss, Layer-Loss) suffer growing accuracy loss as rank decreases when LLaMA-2-7B is quantized to 2-bit, contradicting LoRA's fundamental assumption of low-rank fine-tuning updates. To understand these limitations for 2-bit LLM inference, we investigate quantization error characteristics. Fig.~\ref{fig:w_distance}(b) reveals that weight discrepancy between full-precision and quantized weights increases significantly as bit-precision decreases from 4-bit to 2-bit, with a notable \textit{jump} at 2-bit. This pattern is consistent across weight types and models (LLaMA-2 and LLaMA-3), persisting even with LoftQ adapter initialization. These findings suggest that quantization errors are substantially exacerbated at 2-bit precision, challenging the effectiveness of current LQEC methods.

We further investigate why previous LQEC methods like Weight-SVD of LoftQ (which assume low-rank quantization errors) failed to mitigate the significant discrepancies at 2-bit quantization. While this assumption holds for 3-4-bit quantization, as supported by LQER and SqueezeLLM~\cite{2023.Kim.Keutzer}, its validity for 2-bit quantization remained unexplored. Fig.~\ref{fig:w_distance}(c) illustrates the minimum rank required to suppress weight discrepancy across different bit-precisions. Notably, 3-bit quantization needs only a small adapter rank, but 2-bit quantization demands a much higher rank. This finding suggests that \textit{2-bit quantization errors are inherently high-rank}, challenging the effectiveness of typical SVD-based low-rank adaptation techniques used in existing LQEC methods.

%% file: Sections/methodology.tex
\section{Methodology}
\label{sec:method}

In this section, we first propose the rank sensitivity analysis to reveal the impact of the discrepancy scope on LQEC performance. Based on this intriguing finding, we propose a simple yet effective loss, model-level discrepancy loss (Model-Loss), as a new objective for LQEC that overcomes the rank sensitivity of 2-bit quantization errors.

\begin{figure}[t]
\begin{center}
\centerline{\includegraphics[width=\columnwidth]{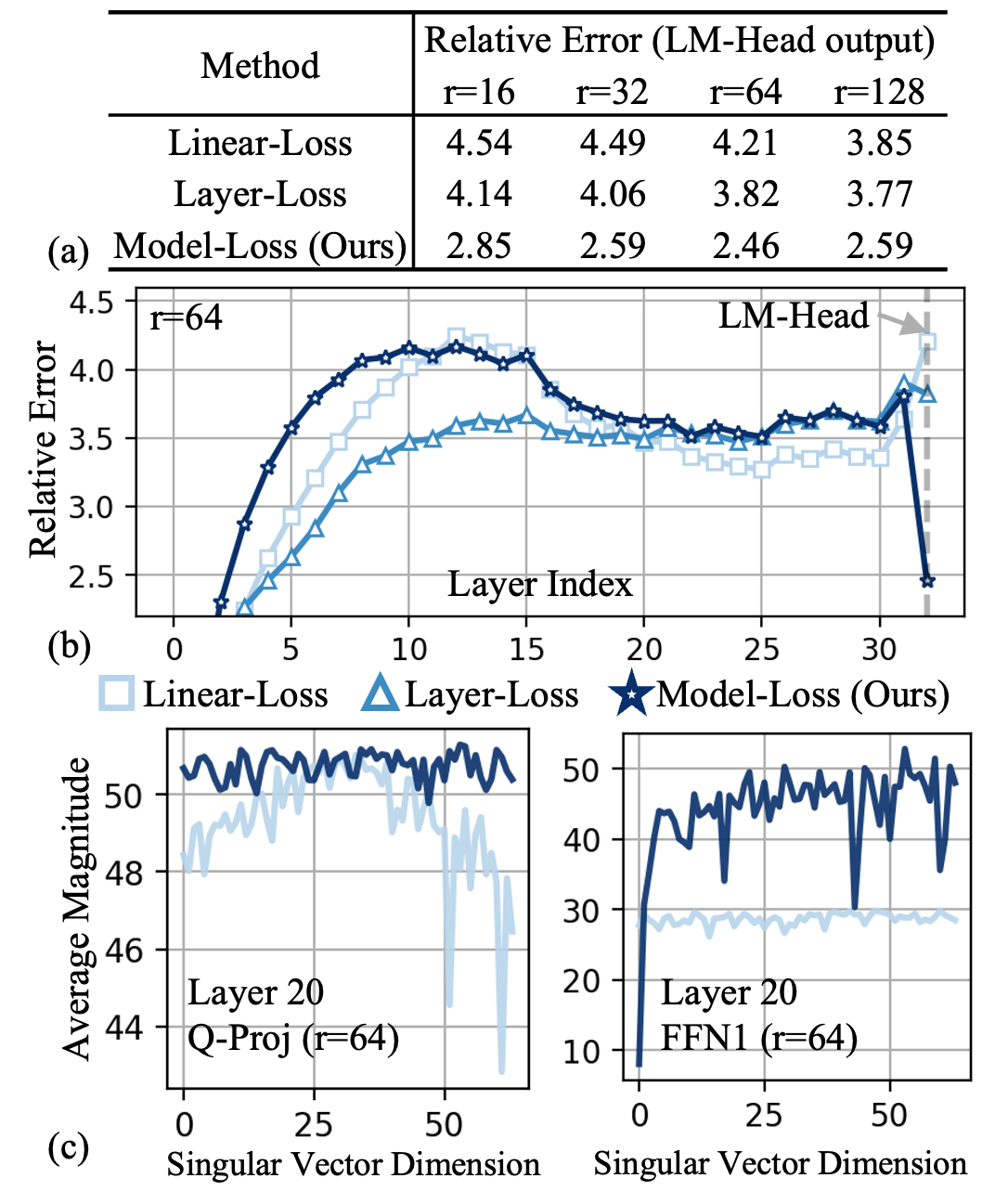}}
\vspace{-3mm}
\caption{(a) Relative error of the LM-head output activation compared to the baseline inference across error compensation strategies, with base weights quantized using OmniQuant. (b) Relative error of intermediate activations and head output compared to baseline inference. (c) Comparison of average magnitudes of left singular vectors between linear and model level optimization.}
\label{fig:granularity}
\vspace{-5mm}
\end{center}
\end{figure}

\subsection{Rank Sensitivity Analysis}
The main limitation of existing LQEC methods is their indiscriminate use of low-rank adapters for QEC without considering quantization error characteristics. To address this, we introduce a new metric called \textit{rank sensitivity}, which measures the relative error ($E=|(Y-Y^q)/Y|$) of logits at the LM-Head output. Lower rank sensitivity (smaller relative error) indicates more accurate inference, as logits directly influence token prediction accuracy. Using this metric, we analyze how the discrepancy scope affects LQEC performance. We extend the discrepancy scope to encompass all Transformer layers, proposing a new discrepancy loss at the output activation of the final ($N$'th) Transformer layer (Model-Loss, Fig.~\ref{fig:optimization}(e)):
\begin{equation}
\underset{L_1, L_2}{\arg\min} \ \|Y_N-Y_N^q\|_{F}.
\label{eq:apiq_obj}
\end{equation}
Fig.~\ref{fig:granularity}(a) compares the rank sensitivity of LQEC methods with varying discrepancy scopes and ranks on LLaMA-2-7B. Two key observations emerge: 1) Rank sensitivity decreases as the discrepancy scope expands from a single linear module to the entire model. 2) With Model-Loss, rank sensitivity remains low even at very small ranks (e.g., rank 16). This suggests that Model-Loss mitigates the high-rank requirements typically associated with 2-bit quantization errors.

To elucidate the rank-insensitive nature of Model-Loss, we compare layer-wise relative errors across three scopes of discrepancy loss in LLaMA-2 (Fig.~\ref{fig:granularity}(b)). Linear-Loss shows higher relative errors at each Transformer layer compared to Layer-Loss, which is expected given the latter's objective of minimizing layer-wise discrepancy. Notably, Model-Loss exhibits even higher relative errors in intermediate layers, but significantly lower error at the LM-Head. This suggests that internal activation drift may facilitate closer alignment of the final activation with error-free full-precision activation, crucial for accurate token generation insensitive to rank. This insight into enhanced signal propagation flexibility aligns with observations in \cite{2023.Kim.Choi} for full-parameter QAT. However, our Model-Loss incorporates low-rank aspects, necessitating a deeper understanding of parameter-efficient error compensation.

We hypothesize that the rank-insensitivity of Model-Loss stems from its wider discrepancy scope, enabling global adapter adjustment to balance compensation between rank-critical and rank-redundant linear modules during LoRA tuning. This concept builds on RA-LoRA's finding that linear modules have varying rank demands for QEC (e.g., Query-projection (Q-Proj) as low-rank, FFN1 as high-rank \cite{ralora}). Unlike RA-LoRA's complex rank adjustment method, Model-Loss implicitly balance this skewed rank sensitivity during adapter tuning. To support this hypothesis, we compare the average magnitude of each element of LoRA’s singular vectors between Linear-Loss and Model-Loss (Fig.~\ref{fig:granularity}(c)). Model-Loss significantly increases the overall magnitudes of FFN1's singular vectors compared to Q-Proj's singular vectors. This suggests that Model-Loss enhances the contribution of singular vectors in rank-critical modules (FFN1) while activating previously idle singular vectors in rank-redundant modules (Q-Proj), fostering cooperative quantization error compensation across Transformer layers. More results corresponding to Fig.~\ref{fig:granularity}(c) are provided in the Appendix.

\subsection{Rank-Insensitive LQEC}
\label{sec:rilq}
Building on insights from the rank-insensitive characteristics of Model-Loss, we introduce \underline{R}ank-\underline{I}nsensitive \underline{L}oRA-based \underline{Q}uantization error compensation (\tech), a novel method for compensating quantization errors in 2-bit LLMs. \tech significantly improves accuracy by implementing Model-Loss and optimizing LoRA adapters in Transformer layers' linear modules using a Model-Loss. As shown in Eq.~\ref{eq:apiq_obj}, \tech uses gradient descent to collectively tune all adapters, minimizing the discrepancy between full-precision and quantized activation outputs ($Y_N-Y_N^q$) of the final layer. This approach effectively addresses inter-weight inconsistencies arising from 2-bit quantization by learning global discrepancy loss from a holistic model perspective. Notably, \tech is particularly advantageous when adapter shapes are constrained to merge with quantized weights for efficient inference (like QA-LoRA), offering a comprehensive solution to LQEC challenges.

To further enhance the language modeling capabilities of LLMs during autoregressive token generation, \tech incorporates a causal language modeling objective with Ground Truth (GT), in the optimization of low-rank adapters (GT-Loss):
\begin{equation}
\underset{\{L_1, L_2\} \in \theta}{\arg\max} \sum_{t=1}^T P\left(x_t \mid x_{<t} ; \theta\right),
\label{eq:apiq_obj2}
\end{equation}
where $x$ represents a token and $T$ the sequence length. While this approach has been utilized in previous QAT methods~\cite{2023.Kim.Choi}, we found it particularly effective in guiding low-rank adapters to improve the model's generation of coherent and contextually appropriate text sequences. This enhancement further aligns 2-bit quantization adjustments with the calibration data. The additional benefits of this method are demonstrated in Table~\ref{table:abl-losstype} through an ablation study, and the entire procedure for \tech is detailed in the Appendix.

%% file: Sections/experiments.tex
\section{Experiments}
\label{sec:experiments}

\input{Tables/unfinetuned_csqa_ppl}

\subsection{Experimental Setup}

\textbf{Tasks and Models}. We evaluate our proposed method for common-sense QA tasks (WinoGrande~\cite{sakaguchi2019winogrande}, PIQA~\cite{bisk2019piqa}, Hellaswag~\cite{zellers-etal-2019-hellaswag}, ARC\_challenge (ARC-C)~\cite{allenai:arc}, ARC\_easy (ARC-E)~\cite{allenai:arc}) and arithmetic reasoning task (GSM8K~\cite{gsm8k}). We focus on two recent open-source pre-trained LLMs, LLaMA-2-7B~\cite{touvron2023llama2} and LLaMA-3-8B~\cite{llama3modelcard}, for evaluating \tech. Additional model scales are in  Table~\ref{table:model_scale} in the ablation study.

\textbf{Quantization Methods}. For a comprehensive performance comparison, we employ \tech alongside state-of-the-art weight quantization techniques, including OmniQuant, QuIP\#, QuaRot, as well as LoftQ (Weight-SVD based LQEC). Each method is sourced from its respective repository\footnote{https://github.com/OpenGVLab/OmniQuant, https://github.com/Cornell-RelaxML/quip-sharp, https://github.com/spcl/QuaRot, https://huggingface.co/LoftQ}. For OmniQuant and QuaRot, we set the group size to 64 (QuIP\# employs a codebook). Other implementation details of each quantization method are specified in the Appendix.

\input{Tables/finetuned_csqa}

\input{Tables/qalora}

\textbf{\tech Implementation Details}. The \tech implementation includes a calibration process for initializing LoRA adapters on quantized models using the C4 dataset (Raffel et al., 2019). Perplexity is evaluated on both the WikiText-2 (Merity et al., 2016) and C4 datasets. For this purpose, a sequence length 512 is employed, and 256 sentences are randomly sampled from the C4 training dataset. During optimization, the Model-Loss and the GT-Loss are applied, with gradient-based optimization performed using the Adam optimizer. The learning rate is fixed at 1e-4, and the batch size is 8. Additional details, including the overall procedure of \tech, the fine-tuning settings, and an analysis of memory costs, are provided in the Appendix.

\input{Tables/rank_sentsitivity_svd}
\input{Tables/3bit-2bit}

\subsection{Experimental Results}

\subsubsection{Direct Error Correction Results.} 
We evaluate the quantization error compensation capability of \tech on LLaMA-2-7B and LLaMA-3-8B for LoftQ and advanced weight quantization techniques. Table~\ref{table:unfinetuned} presents CSQA tasks' accuracy and the perplexity of WikiText-2 and C4. 

Our key observations are as follows:
\begin{itemize}
    \item LoftQ (based on Weight-SVD) is inadequate for 2-bit quantization, as it experiences over 19\% average accuracy loss, making it unsuitable for deployment. On the other hand, advanced weight quantization methods (OmniQuant, QuIP\#, and QuaRot) help recover accuracy.
    \item \tech significantly enhances the accuracy of weight quantization methods. OmniQuant, QuIP\#, and QuaRot experience a 10-19\% average accuracy loss, but \tech recovers this by a significant margin. 
    \item \tech is more effective on LLaMA-3-8B than on LLaMA-2-7B. Existing LoftQ and weight quantization methods suffer greater accuracy loss on LLaMA-3-8B, consistent with recent observations~\cite{huang2024good} that LLaMA-3 is more sensitive to quantization. For example, QuIP\#, the best-performing quantizer in our experiments, experiences higher accuracy degradation on LLaMA-3. In this case, \tech improves the average accuracy of QuIP\# by 8.1\%, a significant boost.
    \item For 3-bit weight quantization, the accuracy improvement by \tech is less noticeable since QuIP\# and other weight quantization methods already approach FP16 accuracy, leaving less room for improvement.
\end{itemize}

\subsubsection{Task-Specific Fine-Tuning Results.}
We further evaluate the task-specific fine-tuning accuracy of \tech on LLaMA-2-7B. LoRA without \tech indicates a default LoRA initialization; one of the adapter pairs is initialized in Gaussian distribution, and the other is zero-initialized. For OmniQuant and QuIP\#, weights are first quantized, and then the LoRA or \tech is further fine-tuned with CSQA and GSM8K task-specific datasets. Table~\ref{table:task-csqa-gsm8k} shows that \tech consistently improves diverse task-specific fine-tuning performances.

\subsubsection{Integrating LoRA into Linear Modules.}
QA-LoRA reduces overhead by merging adapter parameters into quantized weights via shortening the input activation dimension. However, this shortened input dimension can impair error compensation in aggressive quantization scenarios. We demonstrate that \tech's rank-insensitive nature synergistically enhances LQEC performance. To validate \tech within the QA-LoRA framework, we evaluate OmniQuant-quantized models before and after fine-tuning. Table~\ref{table:qalora} shows \tech improves perplexity and CSQA accuracy through error compensation. This enables \tech to match the efficiency of adapter-less quantized LLM inference while significantly boosting accuracy.

\input{Tables/ra_lora}

\subsection{Ablation Study}

\subsubsection{Rank Sensitivity.}
Table~\ref{table:rank_sensitivity} compares \tech with SVD for NormalFloat (NF)~\cite{2023.Dettmers.Zettlemoyer} and OmniQuant methods. 
\tech outperforms SVD at all rank levels in both quantization method, notably achieving better perplexity and accuracy with 16 ranks than SVD with 256 ranks in OmniQuant. This demonstrates \tech's efficiency, delivering superior performance with smaller adapter sizes. Table~\ref{table:23bit-SVD&RILQ} presents the C4 perplexity of SVD and \tech under 2$\mathbin{\sim}$3-bit quantization across varying ranks. We adjust the rank from 16 to 256 and measure the standard deviation ($\sigma$) of the perplexity. With SVD initialization, the perplexity of W3A16 remains stable across different ranks due to the minimal quantization error. However, the reconstruction of the significant 2-bit quantization error is highly sensitive to rank variation, aligning with our observation in Fig.~\ref{fig:w_distance} that higher ranks are required for handling 2-bit errors. In contrast, \tech initialization exhibits rank-insensitive results across both bit widths.

To further validate \tech's rank-insensitive nature, we compare it with standard QA-LoRA (uniform rank for all adapters) and RA-LoRA (rank-adjusted based on sensitivity). As shown in Table~\ref{table:ra_lora}, \tech outperforms both at low rank (rank=16), suggesting effective internal rank management despite using uniform ranks like standard QA-LoRA.

\input{Tables/ablation_loss}

\input{Tables/quip}

\subsubsection{Impact of Scope for Discrepancy Loss.}
Table~\ref{table:abl-losstype} presents an ablation study on LLaMA-2-7B, examining how loss types and optimization granularities affect quantization error compensation. Increasing scope from linear module to model level improves accuracy, highlighting the importance of inter-layer interactions within model. Incorporating GT-Loss with Model-Loss, which provides richer information compared to using either loss individually, further enhances performance, surpassing the effectiveness of GT-Loss alone and serving as an effective optimization guide. This improvement in accuracy aligns with findings of QAT~\cite{2023.Kim.Choi}. The proposed \tech method, which combines these loss objectives at the model level, achieves the best overall performance, highlighting the benefits of global optimization and diverse loss functions for robust error compensation.

\subsubsection{QuIP\# end-to-end FT with \tech.}
We additionally evaluate the cross effects of end-to-end fine-tuning in QuIP\# (QuIP\#-FT) and \tech on LLaMA-3-8B in Table~\ref{table:quip}. As shown in Table~\ref{table:quip}, \tech improves QuIP\#-FT CSQA accuracy by 2\%, highlighting \tech’s ability to enhance fine-tuning beyond standard end-to-end approaches.

\subsubsection{Experiments on Large Models.}
To evaluate the scalability of \tech, we conduct experiments across the LLaMA-2 family, scaling from 7B to 70B parameters. As shown in Table~\ref{table:model_scale}, \tech consistently enhances the perplexity of the quantized models across all sizes, demonstrating the effectiveness of the proposed quantization error compensation technique, even in larger models.

\input{Tables/apdx-scales}

\input{Tables/converge_time}

\subsubsection{Convergence Time of \tech Based on Calibset.}
As shown in Table~\ref{table:converge_time}, \tech with default setting (256 samples and sequence length of 512) converges in under 40 minutes, matching SVD in speed while achieving lower PPL. Extending the calibration sequence length or number of samples can reduce PPL to 9.33, but requires more than 7 hours, supporting our default choice as sufficient.

%% file: Tables/unfinetuned_csqa_ppl.tex
\begin{table*}[ht]
\centering
\renewcommand{\arraystretch}{1.2} 
\resizebox{0.95\textwidth}{!}{%
\begin{tabular}{c|ccc|rrrrrr|rr}
\Xhline{5\arrayrulewidth}
\multirow{2}{*}{Model}      & \multicolumn{1}{c|}{\multirow{2}{*}{Method}}         & \multicolumn{1}{c|}{\multirow{2}{*}{Bit-width}} & \multirow{2}{*}{\tech} & \multicolumn{6}{c|}{Zero-Shot CSQA Accuracy ↑}                                                                                                            & \multicolumn{2}{c}{Perplexity ↓}                       \\ \cline{5-12} 
                            & \multicolumn{1}{c|}{}                                & \multicolumn{1}{c|}{}                           &                       & \multicolumn{1}{c}{WG} & \multicolumn{1}{c}{PIQA} & \multicolumn{1}{c}{HS} & \multicolumn{1}{c}{Arc-c} & \multicolumn{1}{c}{Arc-e} & \multicolumn{1}{c|}{Avg.} & \multicolumn{1}{c}{WikiText2} & \multicolumn{1}{c}{C4} \\ 
                            \Xhline{3\arrayrulewidth} 
\multirow{15}{*}{LLaMA-2-7B} & \multicolumn{3}{c|}{16-bit Baseline}                                                                                           & 69.06                          & 78.07                    & 57.14                     & 43.43                     & 76.30                     & 64.80                     & 5.47                          & 6.97                   \\ \cline{2-12} 
                            & \multicolumn{1}{c|}{\multirow{2}{*}{LoftQ}}          & \multicolumn{1}{c|}{\multirow{8}{*}{W2A16}}     & -                     & 51.14                          & 53.97                    & 27.81                     & 21.25                     & 31.14                     & 37.06                     & 1078.24                       & 728.63                 \\
                            & \multicolumn{1}{c|}{}                                & \multicolumn{1}{c|}{}                           & \checkmark                     & \textbf{57.38}                 & \textbf{66.32}           & \textbf{41.51}            & \textbf{27.47}            & \textbf{55.01}            & \textbf{49.54}            & \textbf{13.28}                & \textbf{14.12}         \\ \cline{2-2} \cline{4-12} 
                            & \multicolumn{1}{c|}{\multirow{2}{*}{OmniQuant}}      & \multicolumn{1}{c|}{}                           & -                     & 59.12                          & 70.18                    & 43.14                     & 28.84                     & 58.12                     & 51.88                     & 11.19                         & 12.42                  \\
                            & \multicolumn{1}{c|}{}                                & \multicolumn{1}{c|}{}                           & \checkmark                     & \textbf{62.83}                 & \textbf{72.47}           & \textbf{46.66}            & \textbf{31.66}            & \textbf{63.97}            & \textbf{55.52}            & \textbf{9.18}                 & \textbf{10.70}         \\ \cline{2-2} \cline{4-12} 
                            & \multicolumn{1}{c|}{\multirow{2}{*}{QuIP\# }} & \multicolumn{1}{c|}{}                           & -                     & 62.51                          & 71.60                    & 43.84                     & 31.48                     & 62.46                     & 54.38                     & 8.90                          & 11.25                  \\
                            & \multicolumn{1}{c|}{}                                & \multicolumn{1}{c|}{}                           & \checkmark                     & \textbf{66.61}                 & \textbf{75.03}           & \textbf{51.68}            & \textbf{37.29}            & \textbf{69.99}            & \textbf{60.12}            & \textbf{6.94}                 & \textbf{8.69}          \\ \cline{2-2} \cline{4-12} 
                            & \multicolumn{1}{c|}{\multirow{2}{*}{QuaRot}}  & \multicolumn{1}{c|}{}                           & -                     & 55.72                          & 62.51                    & 36.42                     & 22.44                     & 49.96                     & 45.41                     & 11.83                         & 19.56                  \\
                            & \multicolumn{1}{c|}{}                                & \multicolumn{1}{c|}{}                           & \checkmark                     & \textbf{62.12}                 & \textbf{72.47}           & \textbf{47.41}            & \textbf{30.38}            & \textbf{64.81}            & \textbf{55.44}            & \textbf{7.57}                 & \textbf{10.21}         \\ \cline{2-12} 
                            & \multicolumn{1}{c|}{\multirow{2}{*}{OmniQuant}}      & \multicolumn{1}{c|}{\multirow{6}{*}{W3A16}}     & -                     & 67.64                          & 77.75                    & 54.97                     & 40.78                     & \textbf{74.49}            & 63.13                     & 6.07                          & 7.54                   \\
                            & \multicolumn{1}{c|}{}                                & \multicolumn{1}{c|}{}                           & \checkmark                     & \textbf{68.35}                 & \textbf{78.02}           & \textbf{55.19}            & \textbf{41.98}            & 74.20                     & \textbf{63.55}            & \textbf{6.02}                 & \textbf{7.51}          \\ \cline{2-2} \cline{4-12} 
                            & \multicolumn{1}{c|}{\multirow{2}{*}{QuIP\# }} & \multicolumn{1}{c|}{}                           & -                     & 67.01                          & 76.22                    & 54.45                     & 40.02                     & 75.13                     & 62.57                     & 6.01                          & 7.60                   \\
                            & \multicolumn{1}{c|}{}                                & \multicolumn{1}{c|}{}                           & \checkmark                     & \textbf{67.64}                 & \textbf{76.93}           & \textbf{55.80}            & \textbf{40.61}            & \textbf{75.76}            & \textbf{63.35}            & \textbf{5.85}                 & \textbf{7.40}          \\ \cline{2-2} \cline{4-12} 
                            & \multicolumn{1}{c|}{\multirow{2}{*}{QuaRot}}  & \multicolumn{1}{c|}{}                           & -                     & 67.72                          & 76.99                    & 54.12                     & 40.78                     & \textbf{74.79}            & 62.88                     & 5.91                          & 7.68                   \\
                            & \multicolumn{1}{c|}{}                                & \multicolumn{1}{c|}{}                           & \checkmark                     & \textbf{67.88}                 & \textbf{77.20}           & \textbf{55.52}            & \textbf{41.81}            & 74.62                     & \textbf{63.41}            & \textbf{5.81}                 & \textbf{7.53}          \\ \hline
\multirow{9}{*}{LLaMA-3-8B}  & \multicolumn{3}{c|}{16-bit Baseline}                                                                                           & 72.61                          & 79.71                    & 60.19                     & 50.43                     & 80.09                     & 68.61                     & 6.14                          & 8.88                   \\ \cline{2-12} 
                            & \multicolumn{1}{c|}{\multirow{2}{*}{LoftQ}}          & \multicolumn{1}{c|}{\multirow{8}{*}{W2A16}}     & -                     & 47.75                          & 53.81                    & 25.91                     & 20.39                     & 26.14                     & 34.80                     & 56168.18                      & 15016.89               \\
                            & \multicolumn{1}{c|}{}                                & \multicolumn{1}{c|}{}                           & \checkmark                     & \textbf{55.64}                 & \textbf{65.02}           & \textbf{37.11}            & \textbf{22.44}            & \textbf{47.31}            & \textbf{45.50}            & \textbf{27.37}                & \textbf{29.22}         \\ \cline{2-2} \cline{4-12} 
                            & \multicolumn{1}{c|}{\multirow{2}{*}{OmniQuant}}      & \multicolumn{1}{c|}{}                           & -                     & 51.85                          & 59.03                    & 32.86                     & 19.28                     & 36.07                     & 39.82                     & 61.79                         & 52.91                  \\
                            & \multicolumn{1}{c|}{}                                & \multicolumn{1}{c|}{}                           & \checkmark                     & \textbf{58.17}                 & \textbf{69.48}           & \textbf{42.66}            & \textbf{27.99}            & \textbf{56.90}            & \textbf{51.04}            & \textbf{17.42}                & \textbf{20.40}         \\ \cline{2-2} \cline{4-12} 
                            & \multicolumn{1}{c|}{\multirow{2}{*}{QuIP\# }} & \multicolumn{1}{c|}{}                           & -                     & 62.35                          & 67.46                    & 44.66                     & 30.80                     & 57.11                     & 52.48                     & 12.74                         & 16.84                  \\
                            & \multicolumn{1}{c|}{}                                & \multicolumn{1}{c|}{}                           & \checkmark                     & \textbf{66.54}                 & \textbf{74.54}           & \textbf{52.48}            & \textbf{38.23}            & \textbf{71.25}            & \textbf{60.61}            & \textbf{9.39}                 & \textbf{12.96}         \\ \cline{2-2} \cline{4-12} 
                            & \multicolumn{1}{c|}{\multirow{2}{*}{QuaRot}}  & \multicolumn{1}{c|}{}                           & -                     & 60.62                          & 65.23                    & 36.77                     & 24.57                     & 57.91                     & 49.02                     & 14.95                         & 27.77                  \\
                            & \multicolumn{1}{c|}{}                                & \multicolumn{1}{c|}{}                           & \checkmark                     & \textbf{68.03}                 & \textbf{73.12}           & \textbf{48.59}            & \textbf{33.70}            & \textbf{68.69}            & \textbf{58.43}            & \textbf{10.13}                & \textbf{15.48}         \\ \Xhline{5\arrayrulewidth}
\end{tabular}%
}
\vspace{-2mm}
\caption{Direct error compensation results. Following the original work, we apply GPTQ~\cite{2023.Frantar.Alistarh} on QuaRot. End-to-end fine-tuning is not applied to QuIP\#. Results for QuIP\# with fine-tuning are presented in Table~\ref{table:quip}.}
\vspace{-3mm}
\label{table:unfinetuned}
\end{table*}

%% file: Tables/finetuned_csqa.tex
\begin{table}[ht]
\centering
\renewcommand{\arraystretch}{1.2}
\resizebox{\columnwidth}{!}{%
\begin{tabular}{cc|ccc|c}
\Xhline{4\arrayrulewidth}
\multicolumn{1}{c|}{\multirow{2}{*}{Method}}    & \multirow{2}{*}{\tech} & \multicolumn{3}{c|}{CSQA Tasks↑}                  & \multirow{2}{*}{GSM8K↑} \\
\multicolumn{1}{c|}{}                           &                       & PIQA           & Arc-c          & Arc-e          &                        \\ \Xhline{3\arrayrulewidth}
\multicolumn{2}{c|}{16-bit LoRA Fine-Tuning}                                                                              & 79.05          & 47.70          & 79.25          & 38.97                      \\ \hline
\multicolumn{1}{c|}{\multirow{2}{*}{OmniQuant}} & -                     & 77.31          & 38.31          & 69.95          & 28.66                  \\
\multicolumn{1}{c|}{}                           & \checkmark             & \textbf{77.91} & \textbf{40.61} & \textbf{71.70} & \textbf{30.86}         \\ \hline
\multicolumn{1}{c|}{\multirow{2}{*}{QuIP\#}}    & -                     & 78.07          & 44.62          & 74.75          & 35.56                  \\
\multicolumn{1}{c|}{}                           & \checkmark             & \textbf{78.45} & \textbf{45.99} & \textbf{76.01} & \textbf{36.32}         \\ \Xhline{4\arrayrulewidth}
\end{tabular}%
}
\vspace{-2mm}
\caption{Task-specific fine-tuning results for CSQA tasks and GSM8K on LLaMA-2-7B.}
\vspace{-3mm}
\label{table:task-csqa-gsm8k}
\end{table}

%% file: Tables/qalora.tex
\begin{table}[ht]
\centering
\renewcommand{\arraystretch}{1.2}
\resizebox{0.9\columnwidth}{!}{%
\begin{tabular}{c|c|crr|c}
\Xhline{4\arrayrulewidth}
\multirow{2}{*}{\begin{tabular}[c]{@{}c@{}}Inference\\ Bit-width\end{tabular}} & \multirow{2}{*}{\tech} & \multicolumn{3}{c|}{Error Compensation} & Fine-Tuning \\
 &  & CSQA↑ & \multicolumn{1}{c}{Wiki2↓} & \multicolumn{1}{c|}{C4↓} & GSM8K↑ \\ \Xhline{2.5\arrayrulewidth}
\multirow{2}{*}{W2A16} & - & \multicolumn{1}{r}{47.42} & 12.92 & 9.23 & 17.89 \\
 & \checkmark & \multicolumn{1}{r}{\textbf{54.51}} & \textbf{10.65} & \textbf{8.14} & \textbf{23.73} \\ \Xhline{4\arrayrulewidth}
\end{tabular}%
}
\vspace{-2mm}
\caption{QA-LoRA implementation of 2-bit LLaMA-2-7B with \tech.  Weights are quantized with OmniQuant (Calibration set: WikiText-2), evaluated with accuracy (CSQA, GSM8K) and perplexity (WikiText-2, C4).}
\vspace{-5mm}
\label{table:qalora}
\end{table}

%% file: Tables/rank_sentsitivity_svd.tex
\begin{table*}[h]
\centering
\renewcommand{\arraystretch}{1.2}
\resizebox{0.95\textwidth}{!}{%
\begin{tabular}{c|c|c|c|rrrrrr|rr}
\Xhline{5\arrayrulewidth}
\multirow{2}{*}{\begin{tabular}[c]{@{}c@{}}Quantization\\ Method\end{tabular}}     & \multirow{2}{*}{Bit-width}    & \multirow{2}{*}{Rank} & \multirow{2}{*}{LQEC} & \multicolumn{6}{c|}{Zero-Shot Accuracy ↑}                                                                                                                      & \multicolumn{2}{c}{Perplexity ↓}                       \\
                            &                         &                       &                             & \multicolumn{1}{c}{WG} & \multicolumn{1}{c}{PIQA} & \multicolumn{1}{c}{HS} & \multicolumn{1}{c}{Arc-c} & \multicolumn{1}{c}{Arc-e} & \multicolumn{1}{c|}{Avg.} & \multicolumn{1}{c}{WikiText2} & \multicolumn{1}{c}{C4} \\ \Xhline{3\arrayrulewidth}
\multirow{10}{*}{\multirow{2}{*}{\begin{tabular}[c]{@{}c@{}}NormalFloat \\ \cite{2023.Dettmers.Zettlemoyer} \end{tabular}}}
    & \multirow{20}{*}{W2A16} & \multirow{2}{*}{16}   & SVD                         & 51.07                  & 55.33                    & 28.19                  & 19.80                     & 30.85                     & 37.05                     & 1311.39                       & 771.04                 \\
                            &                         &                       & \tech                        & \textbf{55.01}         & \textbf{66.59}           & \textbf{39.59}         & \textbf{25.60}            & \textbf{53.32}            & \textbf{48.02}            & \textbf{14.92}                & \textbf{15.34}         \\ \cline{3-12} 
                            &                         & \multirow{2}{*}{32}   & SVD                         & 49.96                  & 53.97                    & 27.12                  & 21.33                     & 29.29                     & 36.33                     & 1379.42                       & 872.47                 \\
                            &                         &                       & \tech                        & \textbf{57.62}         & \textbf{66.10}           & \textbf{39.83}         & \textbf{24.32}            & \textbf{50.55}            & \textbf{46.97}            & \textbf{13.91}                & \textbf{14.79}          \\ \cline{3-12} 
                            &                         & \multirow{2}{*}{64}   & SVD                         & 51.14                  & 53.97                    & 27.81                  & 21.25                     & 31.14                     & 37.06                     & 1078.24                       & 728.63                 \\
                            &                         &                       & \tech                        & \textbf{57.38}         & \textbf{66.32}           & \textbf{41.51}         & \textbf{27.47}            & \textbf{55.01}            & \textbf{49.54}            & \textbf{13.28}                & \textbf{14.12}         \\ \cline{3-12} 
                            &                         & \multirow{2}{*}{128}  & SVD                         & 49.01                  & 56.31                    & 29.82                  & 22.61                     & 33.67                     & 38.28                     & 437.65                        & 347.13                 \\
                            &                         &                       & \tech                        & \textbf{57.46}         & \textbf{69.80}           & \textbf{43.12}         & \textbf{28.24}            & \textbf{56.73}            & \textbf{51.07}            & \textbf{11.59}                & \textbf{12.96}         \\ \cline{3-12} 
                            &                         & \multirow{2}{*}{256}  & SVD                         & 49.25                  & 54.41                    & 28.59                  & 20.82                     & 29.92                     & 36.60                     & 471.68                        & 239.91                 \\
                            &                         &                       & \tech                        & \textbf{60.14}         & \textbf{70.35}           & \textbf{44.65}         & \textbf{29.95}            & \textbf{59.55}            & \textbf{52.93}            & \textbf{10.38}                & \textbf{11.98}         \\ \cline{1-1} \cline{3-12} 
\multirow{10}{*}{OmniQuant} &                         & \multirow{2}{*}{16}   & SVD                         & 59.35                  & 70.13                    & 42.46                  & 27.99                     & 58.21                     & 51.63                     & 11.45                         & 12.71                  \\
                            &                         &                       & \tech                        & \textbf{61.96}         & \textbf{73.01}           & \textbf{46.77}         & \textbf{31.83}            & \textbf{64.14}            & \textbf{55.54}            & \textbf{9.24}                 & \textbf{10.79}         \\ \cline{3-12} 
                            &                         & \multirow{2}{*}{32}   & SVD                         & 58.96                  & 70.35                    & 42.91                  & 28.58                     & 58.42                     & 51.84                     & 11.34                         & 12.50                  \\
                            &                         &                       & \tech                        & \textbf{62.59}         & \textbf{73.07}           & \textbf{47.24}         & \textbf{31.91}            & \textbf{64.10}            & \textbf{55.78}            & \textbf{9.29}                 & \textbf{10.83}         \\ \cline{3-12} 
                            &                         & \multirow{2}{*}{64}   & SVD                         & 58.48                  & 70.62                    & 43.48                  & 28.75                     & 58.71                     & 52.01                     & 10.96                         & 12.19                  \\
                            &                         &                       & \tech                        & \textbf{62.83}         & \textbf{72.47}           & \textbf{46.66}         & \textbf{31.66}            & \textbf{63.97}            & \textbf{55.52}            & \textbf{9.18}                 & \textbf{10.70}         \\ \cline{3-12} 
                            &                         & \multirow{2}{*}{128}  & SVD                         & 59.91                  & 70.62                    & 44.11                  & 29.27                     & 59.72                     & 52.73                     & 10.47                         & 11.70                  \\
                            &                         &                       & \tech                        & \textbf{62.35}         & \textbf{72.85}           & \textbf{46.29}         & \textbf{31.14}            & \textbf{63.89}            & \textbf{55.30}            & \textbf{9.17}                 & \textbf{10.70}         \\ \cline{3-12} 
                            &                         & \multirow{2}{*}{256}  & SVD                         & \textbf{61.80}         & 71.65                    & 45.90                  & \textbf{31.66}            & 62.16                     & 54.63                     & 9.56                          & 10.99                  \\
                            &                         &                       & \tech                        & 61.48                  & \textbf{73.01}           & \textbf{46.31}         & \textbf{31.66}            & \textbf{63.59}            & \textbf{55.21}            & \textbf{9.17}                 & \textbf{10.72}         \\ \Xhline{5\arrayrulewidth}
\end{tabular}%
}
\vspace{-2mm}
\caption{Ablation study on rank sensitivity comparison between SVD and \tech. Accuracy and PPL are measured right after the quantization error compensation, not after task-specific fine-tuning. (NF2 \& SVD = LoftQ) (Model: LLaMA-2-7B).}
\label{table:rank_sensitivity}
\vspace{-3mm}
\end{table*}

%% file: Tables/3bit-2bit.tex
\begin{table}[h]
\centering
\renewcommand{\arraystretch}{1.2} 
\resizebox{\columnwidth}{!}{%
\begin{tabular}{c|c|ccccc|c}
\Xhline{4\arrayrulewidth}
\multirow{2}{*}{LQEC} & \multirow{2}{*}{Bit} & \multicolumn{5}{c|}{Rank} & \multirow{2}{*}{$\sigma$} \\
                      &        & 16 & 32 & 64 & 128 & 256 &                        \\ \Xhline{2.5\arrayrulewidth}
\multirow{2}{*}{SVD}  & W3A16  & 7.56  & 7.55  & 7.53  & 7.50   & 7.45   & \textbf{0.04}   \\
                      & W2A16  & 12.71 & 12.50 & 12.19 & 11.70  & 10.99  & \textit{0.69}   \\ \hline
\multirow{2}{*}{RILQ} & W3A16  & 7.52  & 7.50  & 7.50  & 7.50   & 7.50   & \textbf{0.01}   \\
                      & W2A16  & 10.79 & 10.83 & 10.87 & 10.70  & 10.72  & \textbf{0.07}   \\ 
                      \Xhline{4\arrayrulewidth}
\end{tabular}%
}
\caption{C4 PPL↓ with different LQEC and bit-width (OmniQuant, LLaMA-2-7B).}
\vspace{-5mm}
\label{table:23bit-SVD&RILQ}
\end{table}

%% file: Tables/ra_lora.tex
\begin{table}[t]
\centering
\renewcommand{\arraystretch}{1.2}
\resizebox{0.9\columnwidth}{!}{%
\begin{tabular}{c|cccc}
\Xhline{4\arrayrulewidth}
\multirow{2}{*}{Method} & \multicolumn{4}{c}{Zero-Shot Accuracy ↑} \\

                        & PIQA             & ARC-C           & ARC-E        & Avg.             \\ 

\Xhline{2.5\arrayrulewidth} 
QA-LoRA (Baseline)               & 73.83            & 34.56           & 65.53          & 57.97            \\ \hline

RA-LoRA                 & 74.54            & 36.18           & 66.33          & 59.02            \\ \hline

\tech                   & \textbf{76.39}   & \textbf{36.86}  & \textbf{68.98} & \textbf{60.74}    \\

\Xhline{4\arrayrulewidth}
\end{tabular}%
}
\caption{Comparison of RA-LoRA and \tech under QA-LoRA setting for task-specific fine-tuning for CSQA. RTN is used for 2-bit weight quantization (LLaMA-2-7B, rank=16).}
\label{table:ra_lora}
\end{table}

%% file: Tables/ablation_loss.tex
\begin{table}[t]
\centering
\renewcommand{\arraystretch}{1.2} 
\resizebox{\columnwidth}{!}{%
\begin{tabular}{c|cc|cccccc}
\Xhline{4\arrayrulewidth}
\multirow{2}{*}{Scope} & \multicolumn{2}{c|}{Loss} & \multicolumn{6}{c}{Zero-Shot Accuracy ↑}                                                              \\
                             & Act         & GT          & WG             & PIQA           & HS             & Arc-c          & Arc-e          & Avg.           \\ 
                             \Xhline{2.5\arrayrulewidth}
Linear                       & \checkmark   & -           & 55.56          & 65.29          & 36.58          & 21.50          & 46.17          & 45.02          \\ \hline
Layer                        & \checkmark   & -           & 58.25          & 68.72          & 43.35          & 29.78          & 56.62          & 51.34          \\ \hline
\multirow{3}{*}{Model}       & \checkmark   & -           & 61.17          & 69.59          & 43.74          & 30.29          &                                 59.05          & 52.77          \\
                             & -            & \checkmark   & 62.04 & \textbf{71.16} & 45.13 & 29.95 & 58.12 & 53.28 
                                \\
                             & \checkmark   & \checkmark   & \textbf{62.59} & 70.89 & \textbf{45.51} & \textbf{32.85} & \textbf{61.70} & \textbf{54.71} \\ \Xhline{4\arrayrulewidth}     
\end{tabular}%
}
\caption{Ablation study on expansion of ApiQ with different discrepancy loss scope: GT, Linear, Layer, Model-Loss (GT \& Model-Loss=\tech) (Model: LLaMA-2-7B).}
\label{table:abl-losstype}
\end{table}

%% file: Tables/quip.tex
\begin{table}[t]
\centering
\renewcommand{\arraystretch}{1.2} 
\resizebox{0.9\columnwidth}{!}{%
\begin{tabular}{c|c|c|c|cc}
\Xhline{4\arrayrulewidth}
\multirow{2}{*}{Model}                                                & \multirow{2}{*}{QuIP\# FT} & \multirow{2}{*}{\tech} & \multirow{2}{*}{\begin{tabular}[c]{@{}c@{}}Avg.\\ CSQA↑\end{tabular}} 
& \multicolumn{2}{c}{PPL↓} \\
        &           &           &           & Wiki2         & C4       \\ \Xhline{2.5\arrayrulewidth}
\multirow{4}{*}{\begin{tabular}[c]{@{}c@{}}LLaMA\\ 3-8B\end{tabular}} 
        & -             & -             & 52.5          & 12.7          & 16.8          \\
        & -             & \checkmark    & 60.6          & 9.4           & 13.0          \\
        & \checkmark    & -             & 59.3          & 9.4           & 12.8          \\
        & \checkmark    & \checkmark    & \textbf{61.3} & \textbf{9.1}  & \textbf{12.4} \\ 
        \Xhline{4\arrayrulewidth}
\end{tabular}%
}
\caption{Comparison with QuIP\#-FT (W2A16).}
\vspace{-5mm}
\label{table:quip}
\end{table}

%% file: Tables/apdx-scales.tex
\begin{table}[t]
\centering
\renewcommand{\arraystretch}{1.2} 
\resizebox{0.9\columnwidth}{!}{%
\begin{tabular}{c|c|c|cc}
\Xhline{4\arrayrulewidth}
\multirow{2}{*}{\# Params} & \multirow{2}{*}{Bit-width} & \multirow{2}{*}{\tech} & \multicolumn{2}{c}{Perplexity ↓} \\
                           &                            &                       & WikiText2       & C4             \\ \Xhline{2.5\arrayrulewidth}
\multirow{2}{*}{7B}        & \multirow{6}{*}{W2A16}     & -                     & 1078.24         & 728.63         \\
                           &                            & \checkmark             & \textbf{13.28}  & \textbf{14.12} \\ \cline{1-1} \cline{3-5} 
\multirow{2}{*}{13B}       &                            & -                     & 59.95           & 72.77          \\
                           &                            & \checkmark             & \textbf{9.56}   & \textbf{11.21} \\ \cline{1-1} \cline{3-5} 
\multirow{2}{*}{70B}       &                            & -                     & 12.69           & 16.36          \\ 
                           &                            & \checkmark             & \textbf{6.42}   & \textbf{8.38}  \\ \Xhline{4\arrayrulewidth}
\end{tabular}%
}
\caption{Error compensation results using \tech across different model sizes in the LLaMA-2 families. The quantization method used in this experiment is LoftQ.}
\label{table:model_scale}
\end{table}

%% file: Tables/converge_time.tex
\begin{table}[t]
\centering
\renewcommand{\arraystretch}{1.2} 
\resizebox{0.9\columnwidth}{!}{%
\begin{tabular}{c|c|c|cc|c}
\Xhline{4\arrayrulewidth}
\multirow{2}{*}{LQEC}  & \multirow{2}{*}{\begin{tabular}[c]{@{}c@{}}\# of \\ Samples\end{tabular}} & \multirow{2}{*}{\begin{tabular}[c]{@{}c@{}}Sequence\\ Length\end{tabular}} & \multicolumn{2}{c|}{PPL} & \multirow{2}{*}{Time} \\
                       &              &              & Wiki2          & C4             &                       \\ \Xhline{3\arrayrulewidth}
-                      & -            & -            & 433            & 474            & 0m                    \\ \hline
SVD                    & -            & -            & 176.17         & 221.23         & 31m                   \\ \hline
\multirow{12}{*}{\tech} & 256          & 128          & 11.35          & 13.20           & 10m                   \\
                       & 256          & 256          & 10.39          & 12.25          & 20m                   \\
                       & \textbf{256} & \textbf{512} & \textbf{10.04} & \textbf{11.79} & \textbf{37m}          \\
                       & 256          & 1024         & 9.86           & 11.51          & 1h 39m                \\
                       & 256          & 2048         & 9.61           & 11.25          & 3h 52m                \\
                       & 256          & 4096         & 9.42           & 11.07          & 9h 6m                 \\
                       & 256          & 5120         & 9.46           & 11.17          & 16h 8m                \\
                       
                       & 64           & 512          & 10.91          & 12.78          & 26m                   \\
                       & 128          & 512          & 10.51          & 11.31          & 34m                   \\
                       & 512          & 512          & 9.83           & 11.50           & 1h                    \\
                       & 1024         & 512          & 9.51           & 11.50           & 1h 37m                   \\
                       & 2048         & 512          & 9.34           & 10.98          & 3h 7m                 \\
                       
                       \Xhline{4\arrayrulewidth}
\end{tabular}%
}
\caption{Perplexity and required time for SVD and RILQ (LLaMA-2-7B, 2-bit RTN, rank=16). The default setting is highlighted in bold.}
\vspace{-5mm}
\label{table:converge_time}
\end{table}

%% file: Sections/conclusion.tex
\section{Conclusion}
\label{sec:conclusion}
In this work, we propose \tech, a novel LoRA-based quantization error compensation method that effectively addresses the challenges of 2-bit weight quantization in large language models. By explicitly employing LoRA for quantization error compensation and utilizing a model-wise activation discrepancy loss, \tech enables robust quantization-error compensation while maintaining computational efficiency. Experiments on LLaMA-2 and LLaMA-3 demonstrate the superiority of \tech in improving 2-bit quantized LLM inference accuracy and fine-tuning performance. 

%% file: Sections/appendix.tex
\section*{Appendix}

\subsection{{Procedure of \tech}}
\label{appendix:rilq-procedure}
\tech can be applied in two ways. First, it can be used for direct error correction to restore the base model's generative capability. Alternatively, \tech can serve as initialization method for LoRA, enabling further task-specific fine-tuning. 

\noindent\textbf{Case 1:} Direct Error Correction with \tech

\begin{enumerate}
    \item \textbf{Generate Quantized Model.} Apply an existing PTQ method (e.g., RTN, OmniQuant) to obtain a weight-quantized student model from the full-precision model (teacher).
    \item \textbf{Add LoRA Module.} Freeze both models, and add a trainable LoRA module to each linear layer of the student model.
    \item \textbf{Initialize LoRA Parameters.} Initialize LoRA parameters using gradient descent on Model-Loss (Eq.~\ref{eq:apiq_obj}) and GT-Loss (Eq.~\ref{eq:apiq_obj2}) on small calibration samples.
\end{enumerate}

\textbf{Note:} Step 3 varies with LQEC methods; LoftQ uses SVD while \tech uses gradient descent on Model (+GT)-Loss. The resulting student model with LoRA is then used for inference. 

\noindent\textbf{Case 2:} Task-Specific LoRA Fine-Tuning with \tech

\begin{enumerate}
    \item \textbf{Setup Fine-Tuning.} With \tech-initialized LoRA, freeze student model parameters and make LoRA trainable.
    \item \textbf{Task-Specific Fine-Tuning.} Update the LoRA parameters using gradient descent on GT-Loss with a task-specific dataset.
\end{enumerate}

This approach is similar to Apple's on-device model~\cite{gunter2024apple}, (Sec. 5, accuracy-recovery adapter), but improves on 2-bit LQEC (vs. 3$\mathbin{\sim}$4 bit) with 200$\times$ sample efficiency ($\mathbin{\sim}$50K tokens vs. 10B tokens).

\begin{figure*}[t]
\begin{center}
\centerline{\includegraphics[width=0.9\textwidth]{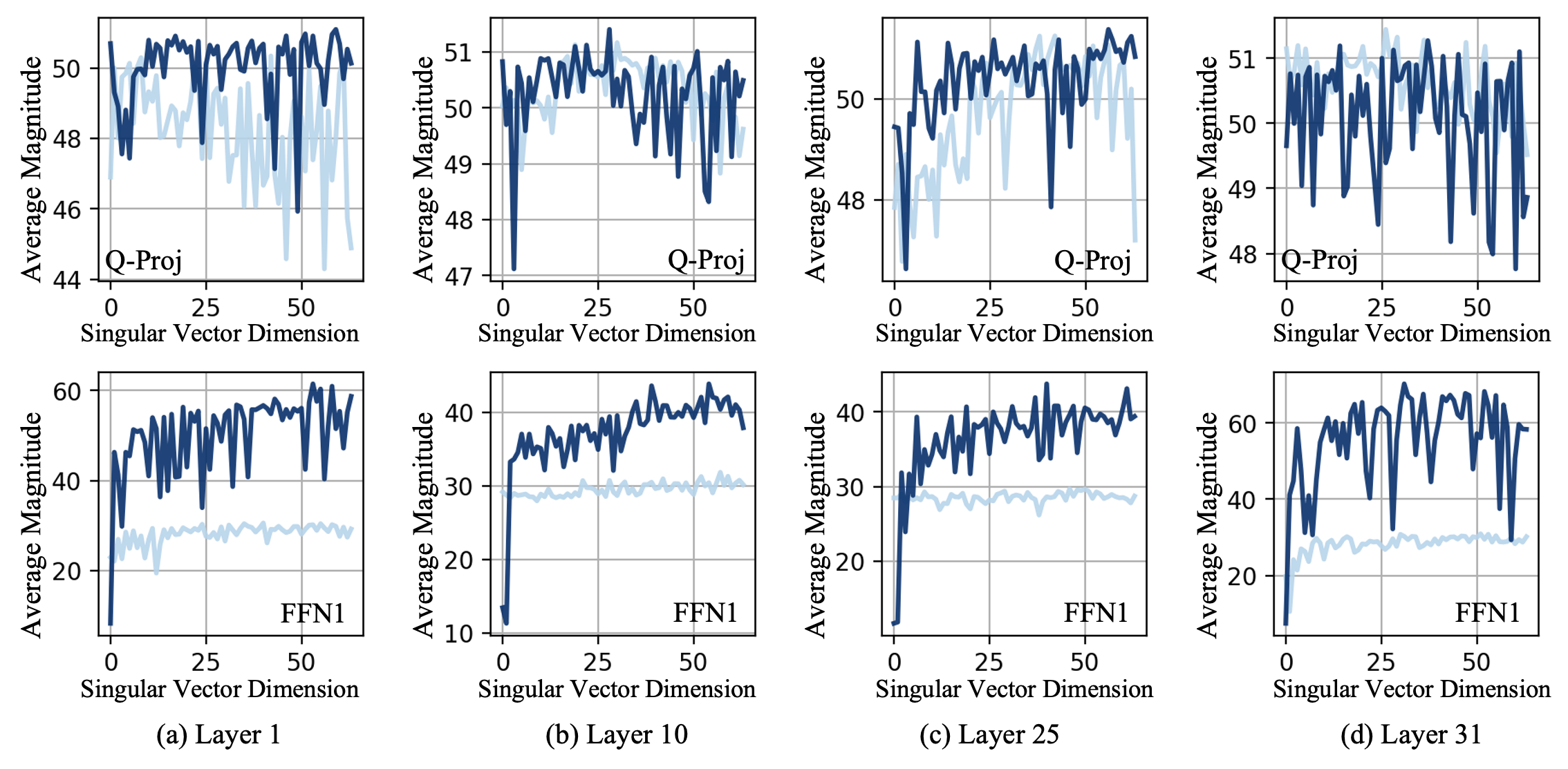}}
\caption{Additional observations on the comparison of the average magnitudes of each element in the left singular vector between linear and model-level optimization (Fig.~\ref{fig:granularity}(c)) across different layers.}
\label{fig:apdx-sv}
\end{center}
\end{figure*}

\subsection{Experimental Details}
\label{appendix:experiments}

\subsubsection{Quantization Settings.}
We quantize all weights of the LLM decoder layer to 2-bit to evaluate how effectively \tech compensates for the residual quantization error via LoRA. Following the original works of four existing quantization methods, we apply their respective quantization techniques to the weights. The details of the quantization for each method are outlined below.
\begin{itemize}
    \item \textbf{LoftQ~\cite{2023.Li.Zhao}.} Following the official GitHub repository of LoftQ\footnote{https://github.com/yxli2123/LoftQ}, we quantize the base weights to 2-bit NormalFloat (NF2)~\cite{2023.Dettmers.Zettlemoyer} with a group size of 64 and initialize LoRA using five iteration steps. We exclude 3-bit quantization results for LoftQ from our evaluation due to the absence of 3-bit quantization support in the official implementation.
    \item \textbf{OmniQuant~\cite{2023.Shao.Luo}.} OmniQuant employs 2-bit integer (INT2) quantization for weights, optimizing scales and zeros through learnable weight clipping with a calibration dataset. In our experiments, we use OmniQuant's official GitHub repository and scripts\footnote{https://github.com/OpenGVLab/OmniQuant}, except for the calibration dataset. We randomly sample 256 sentences from the C4~\cite{c4} training dataset, each containing 512 tokens.
    \item \textbf{QuIP\#~\cite{2024.Tseng.Sa}.} Following the official QuIP\# repository\footnote{https://github.com/Cornell-RelaxML/quip-sharp}, we use non-uniform quantization with a 2-bit codebook based on the 8D $E_8$ lattice. To focus exclusively on the error compensation effectiveness in quantized weights, we do not conduct additional end-to-end fine-tuning, which would otherwise update the weights of LayerNorm and LM-Head after quantization for Table~\ref{table:unfinetuned} and~\ref{table:task-csqa-gsm8k}.
    \item \textbf{QuaRot~\cite{ashkboos2024quarot}.} QuaRot employs the randomized Hadamard transformation to alter the data distribution for suitable quantization. We perform weight-only quantization using 2-bit integer (INT2) with a group size of 64, leveraging the official QuaRot repository\footnote{https://github.com/spcl/QuaRot}.
\end{itemize}

\input{Tables/apdx-hidden-logit}

\subsubsection{More Implementation Details on \tech.} 
We implement a parameter-efficient fine-tuning framework using the Hugging Face PEFT library\footnote{https://github.com/huggingface/peft} for \tech, with the default rank set to 64. All experiments are conducted on NVIDIA A100 and A6000 GPUs. We utilize Model-Loss (Eq.~\ref{eq:apiq_obj}) in conjunction with a causal language modeling loss (Eq.~\ref{eq:apiq_obj2}) to reduce quantization error through LoRA. Both losses receive equal weighting, each being assigned a uniform weight of 0.5. The parameters of LoRA are updated for up to 10,000 steps with early-stopping implemented; the optimization process terminates prematurely if there is no further decrease in loss. For Model-Loss, instead of utilizing logits requiring a large dimension ($\sim$50k vocabulary sizes), we use the output of the final decoder layer with a hidden dimension size (4K in LLaMA-2-7B), with QuaRot as an exception due to its rotation matrix needs. As indicated in Table~\ref{tab:hidden-logit}, this selection does not significantly impact performance.

\subsubsection{Task-Specific Fine-Tuning Settings.} To assess the effectiveness of \tech in task-specific fine-tuning, we employ four key benchmarks: PIQA~\cite{bisk2019piqa}, ARC Challenge~\cite{allenai:arc}, ARC Easy~\cite{allenai:arc}, and GSM8K~\cite{gsm8k}. During the training process, the base model remains frozen, and only the LoRA parameters initialized with \tech are updated. In the fine-tuning process, we train using a causal language modeling loss for each dataset. For all datasets, the source maximum length is set to 384 tokens, and the target maximum length is set to 128 tokens. We use the Adam optimizer with a batch size of 16, and learning rate exploration has been conducted for each dataset (CSQA: 3e-5 to 3e-4, GSM8K: 3e-4 to 1e-3). For GSM8K, training spans six epochs, while for CSQA tasks, the training cycles last three epochs.

\subsubsection{Evaluation Settings.} We utilize the lm-evaluation-harness framework from EleutherAI~\cite{eval-harness} to measure the performance of CSQA tasks and GSM8K. For the perplexity evaluation, we compute the perplexity using the WikiText-2 test dataset and 256 samples from the C4 validation dataset, each with a sequence length of 2048.

\subsection{More Results on Fig.~\ref{fig:granularity}(c) in Other Layers}
\label{appendix:singularvectors}
To more generally observe the phenomenon where low-rank adapters are initialized to exhibit different characteristics depending on the discrepancy scope, we further investigated the phenomenon depicted in Fig.~\ref{fig:granularity}(c) across other layers. As shown in Fig.~\ref{fig:apdx-sv}, the consistent increase in the average size of each element in the singular vectors of FFN1 compared to Q-Proj is also observed in other layers.

\input{Tables/apdx-memory}

\subsection{Memory Cost Analysis}
\tech proposes a method to initialize LoRA while effectively compensating for quantization errors by adding LoRA to quantized weights during fine-tuning. During fine-tuning, as shown in Fig.~\ref{fig:overview}(b), the base weights are frozen in a quantized state, which is more memory-efficient compared to 16-bit LoRA fine-tuning. Table~\ref{tab:apdx-memory} illustrates the memory costs in a task-specific fine-tuning scenario using \tech with the LLaMA-2-7B model. Fine-tuning with a full-precision base model at a rank of 64 requires approximately 14.8GB of GPU memory. In contrast, \tech quantizes the base weights to 2-bit and solely enhances LoRA initialization, thus demonstrating the same memory requirement as QLoRA in fine-tuning, which is only 3.5GB.

\subsection{Comparison with BRECQ}
\label{appendix:brecq}
BRECQ~\cite{li2021brecq} also addresses the impact of quantization loss granularity the in CNN domain, proposing that block-wise updates enhance quantization error reconstruction compared to linear-wise optimization. However, a key difference between BRECQ and our analysis is that BRECQ~\cite{li2021brecq} does not establish the unexpected effectiveness of model-loss for 2-bit LQEC. Instead, BRECQ argues that model-wise reconstruction is less effective than block-wise reconstruction as shown in Table 1 of \cite{li2021brecq}, a point we challenge in this paper. Fig.~\ref{fig:granularity}(b) demonstrates that block-loss results in diverged outputs. The crucial distinction lies in BRECQ's application of compensation to the entire weight matrix through full-rank compensation, adjusting quantization parameters such as rounding and clipping values. In contrast, our approach focuses on low-rank parameter-efficient QEC. Although full-rank adaptation methods (e.g., QAT~\cite{2023.Kim.Choi}) substantially reduce quantization error, they impose prohibitive memory demands. Prior LQEC approaches, such as LoftQ, are more memory-efficient but experience significant accuracy degradation at 2-bit quantization. We are the first to identify that 2-bit LQEC is fundamentally constrained by rank sensitivity in the quantization error, even with BRECQ-like block-loss. Model-loss addresses this limitation by enabling cross-layer compensation, facilitating robust 2-bit LQEC with a rank as low as 16.

%% file: Tables/apdx-hidden-logit.tex
\begin{table}[h]
\centering
\renewcommand{\arraystretch}{1.2} 
\resizebox{0.9\columnwidth}{!}{%
\begin{tabular}{c|cc}
\Xhline{4\arrayrulewidth}
\multirow{2}{*}{Model-Loss Optimization Target} & \multicolumn{2}{c}{Perplexity ↓} \\
                                    & WikiText2 & C4 \\ \hline
Final Layer's Output Activation     & 9.18      & 10.70 \\
Logit                               & 9.28      & 10.88 \\ \Xhline{4\arrayrulewidth}
\end{tabular}%
}
\caption{Perplexity according to the optimization target of the Model-Loss (LLaMA-2-7B, 2-bit OmniQuant).}
\label{tab:hidden-logit}
\end{table}

%% file: Tables/apdx-memory.tex
\begin{table}
\centering
\renewcommand{\arraystretch}{1.2} 
\resizebox{\columnwidth}{!}{%
\begin{tabular}{c|ccccc}
\Xhline{4\arrayrulewidth}
\multirow{2}{*}{\begin{tabular}[c]{@{}c@{}}LoRA\\ Fine-Tuning\\ Method\end{tabular}} & \multicolumn{5}{c}{Required Memory (GB)} \\
 & Weight & \begin{tabular}[c]{@{}c@{}}Weight\\ Grad\end{tabular} & \begin{tabular}[c]{@{}c@{}}Optim\\ States\end{tabular} & Act & Total \\ \hline
FP16 & 13.57 & 0.32 & 0.64 & 0.29 & 14.82 \\
W2A16 QLoRA & 2.23 & 0.32 & 0.64 & 0.29 & \textbf{3.49} \\
W2A16 \tech & 2.23 & 0.32 & 0.64 & 0.29 & \textbf{3.49} \\
\Xhline{4\arrayrulewidth}
\end{tabular}%
}
\caption{Comparison of memory requirements on FP16 LoRA, W2A16 QLoRA and \tech during task-specific fine-tuning (LLaMA-2-7B).}
\label{tab:apdx-memory}
\end{table}